\documentclass[conference]{IEEEtran}

\usepackage{amsmath,amssymb,amsfonts}
\usepackage{algorithmic}
\usepackage{graphicx}
\usepackage{textcomp}
\usepackage[numbers]{natbib}
\usepackage{url}
\usepackage{amsthm}
\usepackage{caption}
\usepackage{subcaption}
\usepackage{balance}
\usepackage{soul}
\usepackage{comment}

\newtheorem{d1}{Definition}

\newtheorem{p1}{Proposition}

\usepackage{multirow}
\usepackage[table,xcdraw]{xcolor}

\definecolor{dkred}{rgb}{0.5, 0.0, 0.13}
\definecolor{comcol}{rgb}{0.01, 0.75, 0.24}
\definecolor{dartmouthgreen}{rgb}{0.05, 0.5, 0.06}

\title{Using Quantifier Elimination to Enhance the Safety Assurance of Deep Neural Networks
}

\author{\IEEEauthorblockN{Hao Ren, Sai Krishnan Chandrasekar, Anitha Murugesan\\}
\IEEEauthorblockA{Honeywell Aerospace, Plymouth, MN \\
{\{hao.ren2,saikrishnan.chandrasekar, anitha.murugesan\}@honeywell.com}}}

\begin{document}

\maketitle

\begin{abstract}
    Advances in the field of Machine Learning and Deep Neural Networks (DNNs) has enabled rapid development of sophisticated and autonomous systems. However, the inherent complexity to rigorously assure the safe operation of such systems hinders their real-world adoption in safety-critical domains such as aerospace and medical devices. Hence, there is a surge in interest to explore the use of advanced mathematical techniques such as formal methods to address this challenge. In fact, the initial results of such efforts are promising. Along these lines, we propose the use of quantifier elimination (QE) --- a formal method technique, as a complimentary technique to the state-of-the-art static analysis and verification procedures. Using an airborne collision avoidance DNN as a case example, we illustrate the use of QE to formulate the precise range forward propagation through a network as well as analyze its robustness. We discuss the initial results of this ongoing work and explore the future possibilities of extending this approach and/or integrating it with other approaches to perform advanced safety assurance of DNNs.

\end{abstract}
 
\section{Introduction}
Recently, there is a tremendous surge of interest within the aerospace community to leverage advances in Machine Learning (ML) to develop sophisticated software for large, autonomous avionic systems such as unmanned aircrafts. In fact, the inherent ability of the modern structurally complex computing systems such as Deep Neural Networks (DNN), that automatically learn and generalize behaviors based on a set of training data rather than explicit programming based on requirements, makes it a natural choice for developing autonomous components for aircraft. However, there is a wide-spread apprehension about deploying such systems in the real-world since it has not been possible to rigorously interpret and assure the safe functional boundaries and behaviors of the DNNs due to their structural complexity and behavioural immensity~\cite{SzegedyZSBEGF13, huang2018safety, alves2018considerations}. For instance, analyzing the robustness of DNNs against \textit{adversarial attacks} \cite{akhtar2018threat,su2019one,yuan2019adversarial} --- small perturbations to inputs that lead to unsafe outputs, remains as an open safety assurance concern.

The use of mathematical techniques such as formal methods has been recently demonstrated as a promising direction towards addressing this challenge. Several formal approaches have been proposed so far to measure DNN's robustness and resilience against adversarial inputs. Namely, recent approaches such as \emph{Reluplex} \cite{ReluplexCav2017,katz2017towards,katz2017reluplex} have used linear programming (LP) and satsifiability modulo theory (SMT) solvers to verify adversarial robustness of DNNs that use piece-wise linear activation function such as Rectified Linear Units (ReLU). Further, \textit{DeepSafe} \cite{gopinath2017deepsafe} studies the safe boundaries of adversarial robustness guided by data, relying on clustering to identify well-defined geometric regions as candidate safe regions. Then it leverages Reluplex verification results for confirmation. Another distinctive approach in the area, \emph{ReluVal} \cite{wang2018formal}, combines the search for concrete counterexamples with layer-by-layer reachability analysis of the DNN. This approach uses interval arithmetic to symbolically compute the range bounds on the intermediate nodes and outputs of the DNN. This has allowed ReluVal to verify a class of DNN properties approximately 200$\times$ faster than Reluplex~\cite{wang2018formal}. However, the underlying algorithm of ReluVal computes the over-approximation of the ranges of nodes in a DNN that, unfortunately, has not been shown to be able to assess the adversarial robustness of the DNN.

In this paper, we propose a \emph{quantifier elimination} (QE) based static analysis of DNNs with ReLU activation, as a complimentary technique to the state-of-the-art DNN verification procedures. QE is a powerful formal technique for gaining insight into problems involving complex logic expressions. 
In the recent past, QE has effectively been used to  derive the strongest system property from components properties in the compositional verification of traditional systems~\cite{ren2018integration,ren2019relic}.
 
Envisioning a DNN as a system composed of layers of connected functional nodes, our idea is to formulate the problem of precise range computation of DNN as a QE problem and leverage QE solvers such as \emph{Redlog}~\cite{dolzmann1997redlog} to derive the range. In addition to property verification, the QE-based analysis allows derivation of precise regions in the input space of a DNN for a desired output robustness measure. We adopt the forward range propagation approach, that has also been used for layer-by-layer reachability analysis in \emph{ReluVal}, to detect linear-behavioral neurons in each layer for simplification of onward computation. The advantage of using QE (over approaches such as ReluVal) is that QE can compute the precise range at the existence of piece-wise linear behavioral neurons, i.e., non-convex scenarios. 

As a proof-of-concept, we developed a prototype implementation and evaluated our approach using pre-trained DNNs of next-generation Airborne Collision Avoidance System for unmanned aircraft (ACAS Xu networks) \cite{julian2016policy} as a case example. To cope with scalability issues we encountered when running the experiments, we used several heuristics and fine-tuning techniques, that we discuss later in the paper. Since ACAS Xu networks have been benchmarks for evaluation of existing formal DNN analysis approaches, in addition to presenting our tests results, we also discuss performance comparison and potentials of methods integration with other tools.

In summary, the contributions of our work are:
\begin{itemize}
    \item QE-based approach for forward range propagation in a DNN.
    \item Initial results of our proof-of-concept implementation and evaluation using ACAS Xu networks.
    \item Commentary on future works of finer-grained implementation for performance elevation.
    \item Commentary on the implications of using QE-based analysis as a complimentary technique to address some of the open assurance-related challenges with DNNs.
\end{itemize}

The rest of the paper is organized as follows. Section~\ref{background} gives the brief background of DNN and related work on current verification progress, and the concept of QE. Section~\ref{QEbased_approach} describes our forward range propagation and verification approach built upon QE formulation of range computation. Section~\ref{case_study} presents our prototype implementation and initial test results. Section~\ref{conclusion} concludes this paper and envisions the potential performance improvement and future work of integrating QE with other formal network verification approaches.

\section{Preliminaries and Background}\label{background}
\subsection{Deep Neural Network}

\begin{figure}[!htb]
\centering
\includegraphics[width=\columnwidth]{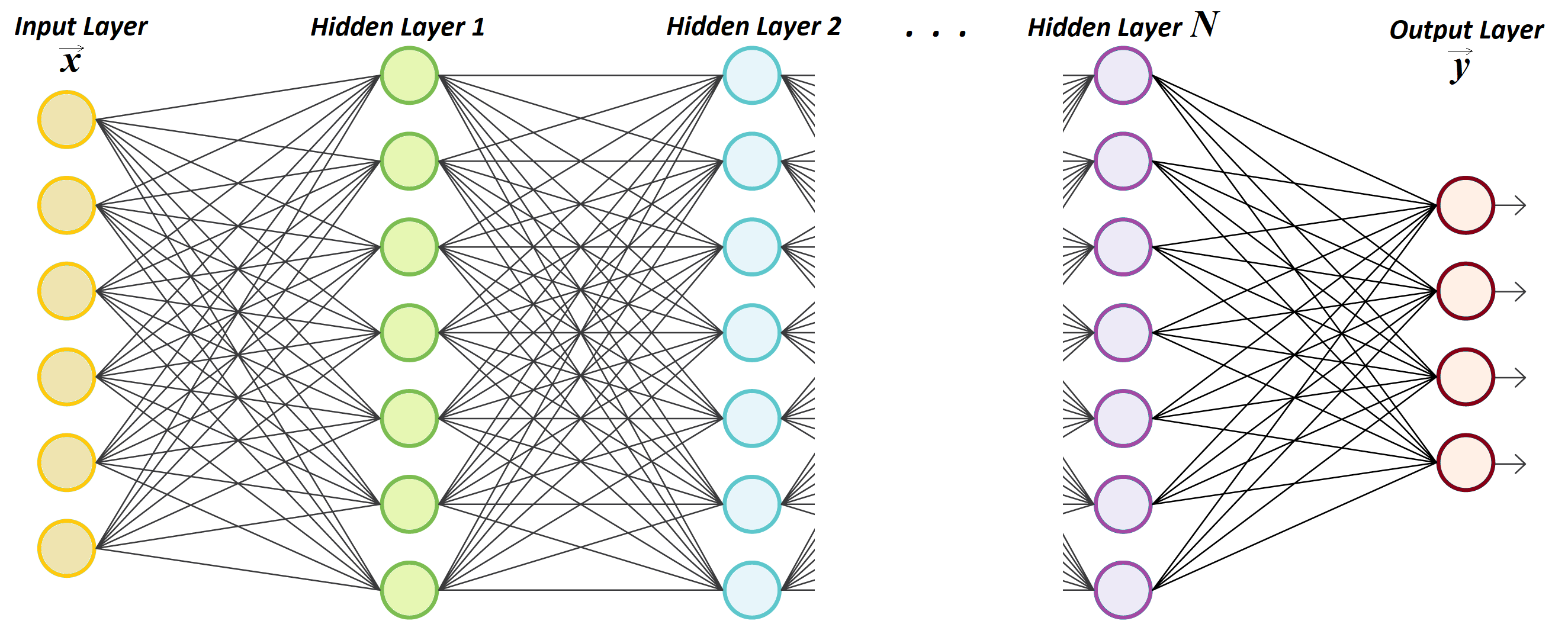}
\caption{N-layer Neural Network}
\label{fig:dnn}
\end{figure}

A typical feed-forward deep neural network is a layered structure of connected nodes called \textit{neurons}, where each neuron refines and extracts information from values computed by neurons in the previous layer. An $N$-layer neural network, as shown in Figure~\ref{fig:dnn}, maps the input vector $\vec{x}$ to output(s) $\vec{y}$, through a composed function $\vec{y}=f(\vec{x}):=f^N(f^{N-1}(\dots f^2(f^1(\vec{x}))\dots))$, where $f^{k\in{[1,N]}}$ denotes the feed-forward computation by the $k^{th}$ layer. Typically, $f^k$ consists of a linear transformation (defined by a weight matrix and a bias vector) of the output values of $f^{k-1}$ ($f^0(\vec{x})=\vec{x}$ for the network input layer), and a non-linear \emph{activation function} to the weighted sum. In this paper, we will consider DNNs with piece-wise linear ReLU-activation function ($ReLU(x)$ = $max(0,x)$ element-wise). 

\subsection{Network Verification} \label{networkVerify}
DNNs are trained on finite data and are then expected to generalize well to previously-unseen inputs, that are similar to the training data. This similarity is loosely defined and hence is met in many domains. However, in safety-critical systems it is desirable, and sometimes mandatory, that it is rigorously assured that certain properties hold irrespective of any input. While the structural complexity of DNNs prevents mathematical interpretation of the DNN's behaviours, it is well-known that testing is insufficient to conclude the non-existence of inputs leading to erroneous results. Hence, there is a strong need to explore alternate means of  analysis procedures such as formal methods that have being extensively used in traditional systems.

Traditionally, formal methods are used to rigorously verify if a desired formalized property holds in a mathematical model of the system. In this paper, we consider the problem of checking whether input-output relationships of a DNN (the mathematical model) hold under a set of input constraints $C_{input}$. The expected output behavior is also described by a set of output constraints $C_{output}$. Verification problem requires showing that the following assertion holds:
\begin{equation*}
    \big(\bigwedge C_{input} \wedge Encoding_{net}\big) \Rightarrow \bigwedge C_{output},
\end{equation*}
where $Encoding_{net}$ is the encoding of inherent constraints of a particular DNN. Typically, $Encoding_{net}$ can be the conjunction of constraints from every neuron, including the linear constraint from the weighted sum function and the piece-wise linear constraint from the ReLU activation function, as exemplified in~\cite{bastani2016measuring}. 

One class of the desired input-output properties of DNNs is robustness against adversarial attacks, i.e., that small input perturbations cause major deviations (such as mis-classification) in the network’s output. A DNN's mis-classification lacks precise mathematical definition, since the ground truth can be subjective. For example, image mis-classification (outside the training data set) is often considered as inconsistency with human eye classification. Human eye, although being close to, rarely qualifies to provide the ground truth. However, adversarial attacks often practically explore and exploit the un-smoothness of a trained network, to carefully craft adversarial inputs. It is due to the fact that DNNs are trained over a finite set of data and are therefore unable to generalize over the entire input space. Dually, a DNN may be retrained to increase its robustness by improving its smoothness \cite{goodfellow2014explaining}. Before we dive into verification, we introduce two common definitions of adversarial robustness below. For a DNN (a classifier) with $n$ neurons at the output layer, each output neuron corresponds to an output label $l\in \{l_1,\dots, l_n\}$. The output label of an input sample $\vec{x}$, denoted by $L(\vec{x})$, is decided based on the comparison among the output neurons' values. For example, one can specify $L(
\vec{x})=l_i$ iff $l_i>l_j, \forall{j_{\in [1,n], j\neq i}}$.

\begin{d1}\label{def1}
A DNN is $\delta$-locally-robust at point 
$\vec{x}_0$ iff
\begin{equation*}
    \forall\vec{x},~~ ||\vec{x}-\vec{x}_0||\le \delta ~~~~\Rightarrow~~~~ L(\vec{x})=L(\vec{x}_0).
\end{equation*}
\end{d1}

Def.~\ref{def1}~\cite{ReluplexCav2017,bastani2016measuring,huang2017safety} requires that the DNN assigns the same label to two input points close enough to each other. Note that, this definition is meaningful only when measured locally regarding the proper choice of the reference point $\vec{x}_0$, since the only type of DNN that satisfies this property globally will naively output the same label for all its inputs (the distance between any two finite input points can be partitioned into finite number of $\delta$ segments). On the other hand, since the DNN classification lacks global ground truth, it is impractical to force a DNN to output the same label within the $\delta$-neighbourhood of $\vec{x}_0$. Therefore, we also consider another definition of DNN robustness that is more inherent to the expected behavior of a DNN itself as follows.

\begin{d1}\label{def2}
A DNN is ($\delta,\epsilon$)-locally-robust at point $\vec{x}_0$ iff
\begin{equation*}
    \forall\vec{x},~||\vec{x}-\vec{x}_0||\le \delta ~\Rightarrow~ \forall f_{\in\{f_1,\dots,f_n\}}, ||f(\vec{x})-f(\vec{x}_0)||\le\epsilon,
\end{equation*}
where $f_{i\in\{1,\dots,n\}}(\vec{x})$ denotes the output value of the $i$-th output neuron given input $\vec{x}$.
\end{d1}

Def.~\ref{def2}~\cite{katz2017towards} addresses the smoothness of the input-output functions locally around a reference input $\vec{x}_0$. Meanwhile, it is also expected to hold for arbitrary reference input $\vec{x}_0$ for some $\epsilon$ (although it is still practically hard to be verified at the global scale even with the state-of-the-art tools).

\subsection{Existing Approaches}
There are usually two directions to approach a verification problem: to falsify a property or prove it. It is sufficient to find one counterexample to falsify a property. Some methods implement searching algorithms for counterexample(s). For example, in convex optimization simplex is an efficient algorithm that moves from one vertex to another along the edges in search for the optimal solution. Reaching one vertex that violates the property leads to algorithm termination along with the vertex identified as a concrete counterexample. In DNNs, the ReLU activation function renders the constraint set non-convex. To address such non-convex optimization problem, Reluplex uses a modified simplex algorithm aided by LP and SMT solvers. During the search, Reluplex either finds a concrete counterexample or concludes that there is none.

On the other hand, proving a property requires to conclude the DNN's behaviors, often realized by reachability analysis that computes the bounds on the values of the (output) neurons. For example, ReluVal uses interval arithmetic to perform forward symbolic interval propagation layer-by-layer. The resulting bounds are \textit{near-precise}, due to the fact that over-approximation has to be used to ``concretize''  the range of ``active'' (branching) neurons for the propagation onward, losing its correlation with other neurons within the same layer. 

It is hard to compare efficiency among verification algorithms just based on the worst-case complexity for two main reasons. Firstly, algorithms aiming for value bounds need to traverse the entire input space for conclusion, regardless of what the property is, whereas when the property is indeed false, falsifying algorithms may terminate at the early stages of the search if they happen to find a counterexample soon. Secondly, the value bounds on neurons, once computed, can be reused for a group of static analysis of the same input constraints. Reusable information shared across verification problems could be a major factor towards scalability enhancement. Therefore, we believe that, just like the training of a DNN usually involves a lot of experimenting and parameter tuning, the verification problems should also be addressed in a broad framework integrating a rich set of approaches and formal methods tools. For a more recent and comprehensive survey of existing algorithms for verifying DNNs, readers can refer to \cite{liu2019algorithms}. To add to that effort, we propose a quantifier elimination (QE) based range propagation method, in the following section, as a complimentary formal method to state-of-the-art verification approaches.

\section{QE-based Approach}\label{QEbased_approach}

\subsection{Quantifier Elimination}\label{QE_intro}
Quantifier elimination (QE) is a powerful technique for gaining insight, through simplification, into problems involving logic expressions in various theories. 
A theory \textit{admits} quantifier elimination if and only if for every quantified formula $\alpha$ in that theory, there exists another quantifier-free formula $\alpha _{QF}$ that is logically equivalent to it. For example, in the domain of real numbers, $\exists x(y>x^2)$ is logically equivalent to $y>0$ (since $x^2$ is always positive). QE is the procedure of deriving an equivalent quantifier-free formula from a quantified formula, as the former can be seen as the residue of the later after the elimination of the quantifiers as well as the quantified variables. In other words, QE can be viewed as a dimensionality reduction method of quantified dimensions. Since it has been proven by Tarski~\cite{tarski1998decision} that the real closed field admits QE, it is suitable for solving the problems of real-world applications.

Following the first implementable QE procedure in 1975 by Collins~\cite{collins1975quantifier}, called the cylindrical algebraic decomposition (CAD), the QE-based techniques and tools have undergone tremendous enrichment over the past few decades and the efforts made along the way have contributed to newer additions. In particular, specialized procedures for restricted problem classes led to newer, more advanced QE procedures, documented in tools such as Mathematica~\cite{reduce} and Redlog~\cite{dolzmann1997redlog}. For example, Redlog implements virtual substitution \cite{weispfenning1988complexity} and partial CAD \cite{collins1991partial} algorithms, that work for formulae where the degrees of the quantified variables are small. 

In our prior work, we have shown that the problem of SMT is essentially an instance of QE. Simply put, checking the satisfiability of a formula $\phi(x_1,\dots,x_n)$ of $n$-variables is equivalent to checking whether or not $\exists x_1\dots\exists x_n ~ \phi(x_1,\dots,x_n)$ can be reduced to $true$ through a QE procedure, since all variables are quantified thereby eliminated, the residue logical formula without variable can only be Boolean value $true$ or $false$. Thus, in the SMT solving domain, the capabilities of SMT solvers and QE solvers overlap in general, yet vary depending on the algorithms they employ and the theories they support. Intuitively, QE, as a more generalized mathematical tool, provides a broad range of functionalities over SMT. For example, while a SMT query returns a satisfying solution to a set of constraints, QE query can apply existential quantifiers on a set of arbitrarily selected variables and return a logical formula regarding only the un-quantified variables revealing the strongest relationship among them. 

As part of our exploration of aforementioned distinguished capabilities of QE, we established in \cite{ren2018integration}, that QE serves as a composition calculus, and applies to compositional verification, a technique being developed to cope with state-space explosion in concurrent systems \cite{stewart2017architectural,henzinger2001assume}. Essentially the strategy of divide-and-conquer is employed where one first establishes the properties of the system components, and then uses those to establish the overarching properties of a complex system.
Supposing a system is composed of $N$ components, the property contract of the $i^{th}$ component can be expressed as $A_i\Rightarrow G_i$, where
$A_i$ (the ``assumption'') and $G_i$ (the ``guarantee'') are both expressed as 1st-order logic formulae over the set of component variables. Then the set of all the system behaviors is constrained by the conjunction of all the components’ contracts, $\bigwedge^N
_{i=1}(A_i\Rightarrow G_i)$. Under these components contracts, the ``strongest system property'' that can be claimed by the system, can be obtained by existentially quantifying the system's internal variables in the conjunct of component contracts with the constraints resulting from the connectivity relation among the components. Thus we established that QE serves as a foundation for property contract composition. Then to check whether a system satisfies a postulated property, we only need to check if the postulated property is implied by the aforementioned strongest system property. This itself can be cast as another QE problem.

\subsection{QE-based DNN Range Propagation}

We extend our idea of contract composition for deriving the strongest system property to formulate the range computation of a ReLU-DNN neuron into a QE problem. Consider a ReLU-DNN  of $N$ layers with input $\vec{x}$, output $\vec{y}$, and label $L$. Let $Z$ be the set of intermediate weighted sum variables and $A$ be the set of the intermediate activation variables. Without loss of generality, we present out approach over the range computation of a single output variable $y\in \vec{y}$. Because all the output variables are structurally parallel and share the same upstream $N-1$ layers, the same range computation technique can be applied to each of them. Let the encoding of DNN regarding to the particular output $y$ (of all its related weighted sum and activation constraints), $Encoding_{net}$, be expressed as the conjunction of all the predicates over $\vec{x}$, $y$, $Z$, and $A$. Let $C_{\vec{x}}$ be the set of specified input constraints (ranges).

As discussed in the previous subsection, we can formulate the strongest property of an output neuron $y$ of a DNN as:
\begin{equation}\label{eq1}
    \exists\vec{x}~\exists Z~\exists A~\Big(\bigwedge C_{\vec{x}}\wedge Encoding_{net}\Big)
\end{equation}
where $\exists$ over a vector (resp. set) denotes existentially quantifying every element variable of the vector (resp. set). Note that, \eqref{eq1} is the DNN interpretation of the strongest system property formulation in \cite{ren2018integration}. Logically, since all the variables but $y$ in \eqref{eq1} are quantified, the quantifier-free equivalence of \eqref{eq1} is a conjunction and/or disjunction of linear predicates over $y$ only. 

\begin{p1}\label{p1}
\eqref{eq1} is the precise range property of $y$ under input range constraints $C_{\vec{x}}$.
\end{p1}
\begin{proof}
Let $R_{precise}(C_{\vec{x}},y)$ denote the actual precise range property of $y$ given $C_{\vec{x}}$. Following Proposition 1 in \cite{ren2018integration}, \eqref{eq1} is a system property of the DNN, i.e., all $y$ values of the DNN under $C_{\vec{x}}$ must satisfy \eqref{eq1}, then we have $R_{precise}(C_{\vec{x}},y)\Rightarrow\eqref{eq1}$. Meanwhile, \eqref{eq1} is the strongest system property, meaning that it implies any other system property. We then also have $\eqref{eq1}\Rightarrow R_{precise}(C_{\vec{x}},y)$. The mutual implication between \eqref{eq1} and $R_{precise}(C_{\vec{x}},y)$ proves that they are in fact equivalent.
\end{proof}

The formulation of \eqref{eq1} and  Proposition~\ref{p1} are the cornerstones of our QE-based DNN range propagation approach. However, $Encoding_{net}$ contains a large size of variables and constraints, especially the piece-wise linear constraints that render the problem non-convex. The state-of-the art QE solvers are insufficient to process it naively.
In the rest of the section, we briefly present a layer-by-layer \emph{forward range propagation} (a sequence of range computations, each of which has its own formulation as in \eqref{eq1}) similar to ReluVal \cite{wang2018formal}, along with a set of heuristics on-the-fly to reduce the size of each individual QE formulation, and eventually conclude the precise range of $y$ for a benchmark-scale network. 

\noindent {\bf Basics of Forward Range Propagation}. In our setting, each hidden neuron is split into a weighted-sum node and an activation node. We denote the output range (resp. value) of the weight sum node as the $z$-range (resp. $z$-value), and the output range (resp. value) of the activation node as the $a$-range (resp. $a$-value). The forward range propagation starts from the first hidden layer, where the $z$-range of each neuron can be formulated similarly to (1) with the smallest sub-graph encoding including only the input constraints and the weighted sum node constraint. Since all neurons in the same layer share the same upstream sub-graph, their $z$-ranges naturally fit multi-threaded \textit{parallel} computation. Theoretically, the time cost for range propagation of an entire layer with $n$ neurons can be reduced to that of a single neuron by running $n$ threads in parallel. The $a$-range of a neuron can be trivially computed from its $z$-range following the ReLU function. Such computation steps are forward progressed layer-by-layer until the ranges of the target output neurons are computed.

\noindent {\bf On-the-fly Pruning for Scalability}. The scalability of the precise range propagation largely depends the QE solver's performance on the encoding of the network. The classes of local adversarial robustness properties verification have relatively small input ranges ($\delta$-neighborhood of a sample point). In this premise, a majority of the hidden neurons either behave purely linearly or even have a constant zero output. The above forward propagation approach identifies such neurons and simplifies the network \emph{behavioral structure} as much as possible. In our preliminary tests, we found that the major factor causing our QE solver Redlog timeout, is the number of disjunctive predicate clauses in the network encoding. This corresponds to the number of branching neurons remaining in the simplified network behavioral structure. We observed that Redlog can handle a single neuron precise range computation, with 6$\sim$8 branching neurons distributed across different upstream layers, in less than a few tens of seconds. However, one more such neuron may suddenly leads to more than 1,000 seconds execution time or even timeout. The on-the-fly pruning (identifying and then simplifying) of the linear part of the network behavioral structure results in a much computationally affordable residual of the network. Note that, the precise range computation of each neuron in layer $l$ requires the encoding of \textit{all} its upstream sub-graph of the behavioral structure rather than just the ranges of neurons in layer $l-1$. Failure to do so will lose tracking the correlation among the neurons, thus introducing large over-approximation errors similar to the naive interval propagation noted by ReluVal in \cite{wang2018formal}. 

\noindent {\bf Over-approximation}. Many verification problems allow over-approximation as long as the property still holds with a reasonable over-approximation precision. In network node range computation, one can replace the set of constraints of a branching neuron, i.e., constraints of both upstream and its functionality, with \textit{only} its propagated $a$-range to over-approximate the downstream range computation. This effect is known as the dependency problem \cite{moore2009introduction} and is studied in \cite{wang2018formal} in the network context. A new effect for our QE-based range propagation is that the over-approximation may cause more branching neurons downstream than there actually exists. We propose a few techniques to fine-tune the trade-off between computation efficiency and precision. The over-approximation shall be used only for the branching neurons, since branching behavior is the major hinder of QE solving. And/Or one can over-approximate a subset of the branching neurons, leaving a small number of branching neurons as is during the propagation. Another way to reduce over-approximation errors is input partition discussed separately below.

\noindent {\bf Input partition}. With input partition (also adopted by ReluVal in \cite{wang2018formal} in the form of input bisection), the overall range will be the union of range from each input sub-space. Input partition can significantly reduce the accumulated over-approximation error. Meanwhile it suppresses the number of branching neurons in each input sub-space, improving the scalability for both precise and over-approximated forward range propagation. Note that, input partition is highly parameterizable as the partition can be done in one or more input dimensions, in the form of bisection or more segments, and evenly or by decision-boundary. One future research direction is to decide the partition plan possibly depending on input sensitivity analysis. Input partition is also highly parallelizable as the sub-spaces can be checked independently.

\section{Implementation and Case Study}\label{case_study}
We have developed a proof-of-concept prototype implementation of our QE-based forward range propagation and verification framework, with on-the-fly network pruning using propagated neuron ranges. Partials of the proposed over-approximation and input partition methods are used for preliminary tests. 

\subsection{Experimental Set-up}

Our prototype tool is implemented in Python with multi-threaded parallel computation feature. We present our tests results on the benchmark of pre-trained DNNs of a next-generation Airborne Collision Avoidance System for unmanned aircraft (ACAS Xu networks) open-sourced by the original authors \cite{ReluplexCav2017}. We adapted their custom neural network format as per the requirements of our tool. As a first pass, we set a hard threshold of 7200 seconds as our maximum execution time before we declared the test ``timeout (TO)''. All experiments were run on 2 Intel Xeon Silver 4114 CPUs with 40 logical cores and 32 GB memory. All experiments are \emph{10-threaded}. 

\subsection{Overview of ACAS Xu Networks}
\begin{figure}[!htb]
\centering
\includegraphics[width=0.3\textwidth]{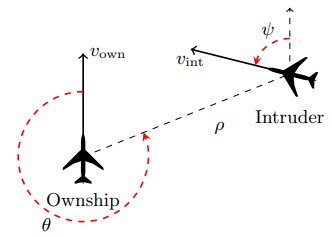}
\caption{Geometry for ACAS Xu Horizontal Logic Table. Figure taken from \cite{ReluplexCav2017}.}
\label{fig:ACASXu}
\end{figure}

ACAS X is a family of aircraft collision avoidance systems, which is currently under development by FAA and NASA \cite{ACASXu}. ACAS Xu system is the version for unmanned aircraft control. It is intended to receive sensor information regarding the drone (the \emph{ownship}) and any nearby \emph{intruder} drones, and then issue horizontal turning advisories aimed at preventing collisions. Each advisory is assigned a score, with the lowest score corresponding to the best action. The input sensor data, as illustrated in Figure~\ref{fig:ACASXu}, are:
\begin{itemize}
\item $\rho$: distance from ownship to intruder;
\item $\theta$: angle of intruder relative to ownship heading direction;
\item $\psi$: heading angle of intruder relative to ownship heading direction;
\item $v_{\text{own}}$: speed of ownship;
\item $v_{\text{int}}$: speed of intruder;
\item $\tau$: time until loss of vertical separation; and
\item $a_{\text{prev}}$: previous advisory.
\end{itemize}

\begin{center}
\begin{table*}[htb]
\begin{tabular}{|l|c|c|c|c|c|c|c|c|c|}
\hline
\rowcolor[HTML]{C0C0C0} 
\multicolumn{1}{|c|}{{ }}                                              & \multicolumn{3}{c|}{{ \textbf{$\delta$=0.005}}}                                                                      & \multicolumn{3}{c|}{{ \textbf{$\delta$=0.01}}}                                                                           & \multicolumn{3}{c|}{{ \textbf{$\delta$=0.025}}}                                                                          \\ \cline{2-10} 
\rowcolor[HTML]{C0C0C0} 
\multicolumn{1}{|c|}{\multirow{-2}{*}{{ \textbf{Tests on DNN: $N_{1,1}$}}}} & { \textbf{~~result~~}}                               & { \textbf{~precise~}} & { \textbf{time}} & { \textbf{result}}                                   & { \textbf{~precise~}} & { \textbf{time}} & { \textbf{result}}                                   & { \textbf{~precise~}} & { \textbf{time}} \\ \hline

{ }                                                                    & { }                      & { no}                   & { 23}            & { }                          & { no}                   & { 24}            & { }                          & { no}                   & { 542}           \\ \cline{3-4} \cline{6-7} \cline{9-10} 

\multirow{-2}{*}{{ \textbf{P1: {[}0, 0, 0, 0, 0{]}}}}                  & \multirow{-2}{*}{{ COC}} & { yes}                  & { TO}            & \multirow{-2}{*}{{ UNKNOWN}} & { yes}                  & { TO}            & \multirow{-2}{*}{{ UNKNOWN}} & { yes}                  & { TO}            \\ \hline

{ }                                                                    & { }                      & { no}                   & { 26}            & { }                          & { no}                   & { 31}            & { }                          & { no}                   & { 85}            \\ \cline{3-4} \cline{6-7} \cline{9-10} 

\multirow{-2}{*}{{ \textbf{P2: {[}0.2, -0.1, 0, -0.3, 0.4{]}}}}        & \multirow{-2}{*}{{ COC}} & { yes}                  & { 112}           & \multirow{-2}{*}{{ COC}}     & { yes}                  & { TO}            & \multirow{-2}{*}{{ UNKNOWN}} & { yes}                  & { TO}            \\ \hline

{ }                                                                    & { }                      & { no}                   & { 16}            & { }                          & { no}                   & { 35}            & { }                          & { no}                   & { 49}            \\ \cline{3-4} \cline{6-7} \cline{9-10} 

\multirow{-2}{*}{{ \textbf{P3: {[}0.45, -0.23, -0.4, 0.12, 0.33{]}}}}  & \multirow{-2}{*}{{ COC}} & { yes}                  & { 21}            & \multirow{-2}{*}{{ COC}}     & { yes}                  & { 75}            & \multirow{-2}{*}{{ COC}}     & { yes}                  & { TO}            \\ \hline

{ }                                                                    & { }                      & { no}                   & { 20}            & { }                          & { no}                   & { 65}            & { }                          & { no}                   & { 287}           \\ \cline{3-4} \cline{6-7} \cline{9-10} 

\multirow{-2}{*}{{ \textbf{P4: {[}-0.2, -0.25, -0.5, -0.3, -0.44{]}}}} & \multirow{-2}{*}{{ COC}} & { yes}                  & { 35}            & \multirow{-2}{*}{{ COC}}     & { yes}                  & { TO}            & \multirow{-2}{*}{{ UNKNOWN}} & { yes}                  & { TO}            \\ \hline

{ }                                                                    & { }                      & { no}                   & { 22}            & { }                          & { no}                   & { 40}            & { }                          & { no}                   & { 143}           \\ \cline{3-4} \cline{6-7} \cline{9-10} 

\multirow{-2}{*}{{ \textbf{P5: {[}0.61, 0.36, 0.0, 0.0, -0.24{]}}}}    & \multirow{-2}{*}{{ COC}} & { yes}                  & { 75}            & \multirow{-2}{*}{{ COC}}     & { yes}                  & { TO}            & \multirow{-2}{*}{{ UNKNOWN}} & { yes}                  & { TO}            \\ \hline
\end{tabular}

\vspace{0.15in}

\footnotesize{\textit{Note: All sample points are in format $[\delta,\theta,\psi,v_{own},v_{int}]$. All times are in seconds. ``TO'' denotes timeout. value in the ``precise'' column denotes if the precise range is propagated or the over-approximation otherwise.}}
\vspace{0.1in}
\caption{$\delta$-local-robustness tests results.}
\label{delta_local_robust}
\end{table*}
\end{center}

ACAS Xu networks \cite{julian2016policy} are the DNN re-implementation of the ACAS Xu system, with the intention to reduce memory constraint of the on-board avionics hardware \cite{ReluplexCav2017,julian2018deep}. In one of the recent versions, the input state space of the ACAS Xu system is partitioned into 45 sub-spaces by the combination of $a_{\text{prev}}$ (5 values) and discretized $\tau$ (9 values) \cite{ReluplexCav2017}. A unique DNN is trained for each sub-space with the remaining 5 inputs. The complete ACAS Xu networks contains an array of 45 such fully-connected DNNs, each of which (denoted as $N_{i_{\in[1,5]},j_{\in[1,9]}}$) has total 300 hidden neurons evenly distributed in 6 hidden layers. Each hidden neuron consists of a linear weighted sum function feeding to a ReLU activation function. Recall that, in this paper, we denote the output range (resp. value) of the weight sum node of a neuron as its $z$-range (resp. $z$-value), and the output range (resp. value) of the activation node as its $a$-range (resp. $a$-value). The output layer of each DNN consists of five output neurons computing and outputting only weighted sum of the $a$-values of neurons in the previous layer, i.e, the output neurons do not possess activation function. The five output neurons are denoted by $Q_{COC}, Q_{WL}, Q_{WR}, Q_{SL}$, and  $Q_{SR}$ respectively, one
for each of the five possible advisories: Clear-of-Conflict (COC), weak left (WL), weak right (WR), strong left (SL), and strong right (SR). An advisory is selected as the DNN output if the corresponding output neuron has the minimal $z$-value.

\subsection{Results}\label{test_results}
In this section, we first present our tests results regarding local adversarial robustness defined by Def~.\ref{def1} in Section \ref{networkVerify}. Then, we introduce our new feature of precise quantitative analysis of the local robustness region regarding Def.~\ref{def2} in Section \ref{networkVerify}. Finally, we show from one test case that our approach has the potential to be used for more general type of DNN input-output property verification.\\

\noindent {\bf $\delta$-Local-Robustness Tests}: $\delta$-local-robustness requires a $\delta$-neighborhood of a sample point agrees on the output label. This is the human basis for identifying mis-classification. We tested on one of the 45 networks, with the same 5 (normalized) sample points by Reluplex \cite{ReluplexCav2017} and 3 perturbation values on the normalized inputs. After forward range propagation, one can compare the range bounds on each output neuron to decide if one of the labels always corresponds to the least value. Table.~\ref{delta_local_robust} shows this verification results based naively on output ranges.

 A few comments on the $\delta$-local-robustness tests results and possible actions for improvement are listed below: 
\begin{itemize}  
    \setlength{\itemsep}{1.3pt}
    \setlength{\parskip}{1.3pt}
    \item The precise range propagation can be achieved at a small perturbation ($\delta$ = 0.005). Note that, the $\delta$ values are perturbations on normalized values, their raw-sizes are still large for reasoning about adversarial robustness.
    \item When $\delta$ increases to 0.01, most precise range propagation causes timeout due to the increasing number of branching neurons. But verification on 4 out 5 sample points can still be completed conclusively using over-approximation. 
    \item The verification result is ``UNKNOWN'' using naive range comparison among output neurons when there is overlapping on those ranges. Input partition (not yet used for this group of tests) can be adopted to refine the range bounds on input sub-space for decision, and/or the range computation formulation can be augmented with extra linear constraints of encoding the guarantee on the correct label getting the minimum value. 
    \item Our timeout threshold is relatively low of 7200 seconds. Some of the ``TOs'' in Table.~\ref{delta_local_robust} actually get results back in longer time. For example, precise forward range propagation of P5 with $\delta=0.01$ on DNN $N_{1,1}$ is completed in 16216 seconds. Since its computation time at $\delta=0.005$ (75 seconds) is drastically lower, it suggests that bisection could improve the case. We will verify that in the near future.
\end{itemize}

\noindent {\bf ($\delta,\epsilon$)-Local-Robustness Tests}: ($\delta,\epsilon$)-local-robustness emphasizes the $\epsilon$-smoothness in the continuous output space (instead of discrete labels) within the $\delta$-neighborhood of a sample point. The key challenge here is to find the precise mapping between the two parameters $\delta$ and $\epsilon$ in dual directions. To the best of our knowledge, our prototype is the first to achieve such precise mappings at the presence of branching neurons. Without loss of generality, we demonstrate our approach on the COC output neuron only.
\subsubsection{$\delta$-to-$\epsilon$ precise mapping}
This precise mapping is a natural result of our QE-based precise forward range propagation given a small $\delta$. For example, the precise range of the $Q_{COC}$ output of DNN $N_{1,1}$ with in the $0.01$-neighborhood of P3 in Table~\ref{delta_local_robust}, is computed to be $[-0.02158274,-0.02157095]$ (rounded to the nearest $10^{-8}$). The largest absolute difference from the sample output of P3 ($-0.02157623$) is the precise $\epsilon$ (=$0.00000651$). 

\subsubsection{$\epsilon$-to-$\delta$ precise mapping} Narrowing down the $\delta$ for a given $\epsilon$ helps isolating adversarial ranges of inputs from the non-adversarial ones. It allows a DNN developer to do so potentially with a desired precision. Both Reluplex and ReluVal use iterative bisections on the original $\delta$ to approach a better precision. Our approach takes this to a further level, allowing direct backward derivation of the precise input perturbation for a desired $\epsilon$. 

Given $\vec{x}_0$ as the sample input and an original reference $\delta_0$, we first perform precise forward range propagation to obtain the corresponding $\epsilon_0$ along with a simplified behavioral structure. 
Then for a desired output deviation $\epsilon^*<\epsilon_0$, we are able to formulate the derivation of the corresponding precise input perturbation $\delta^*$ using QE in a mutated version of \eqref{eq1} with $\delta$ as integrated variable and all the other variables properly quantified. After the QE procedure eliminates all the quantifiers and the quantified variables, its quantifier-free equivalence is a simple predicate with $\delta$ as the only remaining variable. Taking again DNN $N_{1,1}$ and P3 as an example, if we pick the desired $\epsilon^*=0.000003$ ($<0.00000651$), i.e., $\epsilon^*$ is less than the deviation under the original $\delta_0=0.01$, then $\delta\le0.00460093247$ is \emph{derived in a single step}. Hence the precise $\delta^*$ is $0.00460093247$. The time cost for this computation is only 4.5 seconds.\\

\noindent {\bf Remark}: \textit{Before the derivation of $\delta^*$, a forward range propagation with an original $\delta_0$ is necessary to obtain the much simpler network behavioral structure. Otherwise QE solving using the encoding of the entire network is too computational costly to be tractable. Nevertheless, the forward range propagation can be relaxed to allow over-approximation in some cases. When the $\epsilon_0$ computed from $\delta_0$ is an over-approximated value, then the precise $\delta^*$ derived from the desired $\epsilon^*<\epsilon_0$ is valid if and only if $\delta^*\le\delta_0$. Otherwise, the network is not guaranteed to behave under the behavioral structure of $\delta_0$, thus the derivation loses soundness.}

\noindent {\bf A General Property Test}. In addition to local robustness tests presented above, we tested our prototype on one of the original properties, namely,  $\phi_1$ described in Reluplex paper \cite{ReluplexCav2017}. $\phi_1$ has a much larger input space comparing to all the local robustness test cases, and it specifies that the COC output value never exceeds 1500. Our prototype is still under refinement, we are able to test $\phi_1$ on DNN $N_{1,1}$, mainly to evaluate the potential performance improvement with over-approximation and input partition for verification problems. Using input bisection on all five input dimensions, the union of the COC output range of all 32 sub-spaces is $[-522.12258186,
1453.19412714]$ (rounded to the nearest $10^{-8}$). Thus $\phi_1$ is proven to be valid. The average time for each input sub-space range propagation is 82 seconds.
If our prototype has similar performance on the rest of the 45 ACAS Xu networks, and computation resources allow parallel computation of all input sub-spaces, it will be a drastic performance elevation (assumed 2,624 seconds on all 45 networks) comparing to Reluplex ($>$443,560.73 seconds) and ReluVal (14,603.27 seconds).

\section{Conclusion and Future Work}\label{conclusion}

\noindent {\bf Conclusion}. In the paper, we presented our  contribution towards bridging the gap between the advances in DNN applications and the rigorous verification needs arising from safety concerns, in the form of a prototype tool with a proof-of-concept implementation. Our QE-based approach applies to ReLU-activated DNNs. The preliminary tests results are promising and show the potential of state-of-the-art performance with further refined implementation of over-approximation and input partition techniques. A unique feature of our approach is that it is able to forward propagate the neurons ranges precisely at the existence of a few branching neurons, allowing a more fine-grained quantitative adversarial analysis. A second unique feature is its ability to derive the precise input perturbation from a desired output deviation, assisting the DNN developer to isolate and avoid adversarial ranges with a given precision.

\noindent {\bf Future Work}. We plan to focus on 1) refining our over-approximation and input partition algorithms, 2) customizing QE algorithms/solvers to achieve better performance on the classes of QE problems formulated in our approach, and 3) conducting thorough tests to get a comprehensive evaluation and comparison with state-of-the-art tools such as Reluplex and ReluVal. We envision, in the long term, QE-based range propagation can be integrated with other formal verification procedures, to achieve more flexibility and efficiency on a broader range of DNN formal analysis.

\balance

\bibliographystyle{ieeetr}
\bibliography{references}

\end{document}